\documentclass{article}

\usepackage{arxiv}

\usepackage[utf8]{inputenc} % allow utf-8 input
\usepackage[T1]{fontenc}    % use 8-bit T1 fonts
\usepackage{hyperref}       % hyperlinks
\usepackage{url}            % simple URL typesetting
\usepackage{booktabs}       % professional-quality tables
\usepackage{amsfonts}       % blackboard math symbols
\usepackage{nicefrac}       % compact symbols for 1/2, etc.
\usepackage{microtype}      % microtypography
\usepackage{lipsum}
\usepackage{graphicx}
\graphicspath{ {./images/} }
\usepackage{array}
\usepackage{amsmath}
\usepackage{float}
\usepackage{subcaption}
\usepackage{multirow}

\title{Comparative Analysis of the Hidden Markov Model and LSTM: A Simulative Approach}

\author{
  Manie Tadayon\\
  Department of Electrical and Computer Engineering\\
  UCLA University\\
  Los Angeles, CA  \\
  \texttt{manitadayon@ucla.edu} \\
   \And
  Greg Pottie \\
  Department of Electrical and Computer Engineering\\
  UCLA University\\
  Los Angeles, CA \\
  \texttt{pottie@ee.ucla.edu} \\
}
\begin{document}
\maketitle
\begin{abstract}
Time series and sequential data have gained significant attention recently since many real-world processes in various domains such as finance, education, biology, and engineering can be modeled as time series. Although many algorithms and methods such as the Kalman filter, hidden Markov model, and long short term memory (LSTM) are proposed to make inferences and predictions for the data, their usage significantly depends on the application, type of the problem, available data, and sufficient accuracy or loss. In this paper, we compare the supervised and unsupervised hidden Markov model to LSTM in terms of the amount of data needed for training, complexity, and forecasting accuracy. Moreover, we propose various techniques to discretize the observations and convert the problem to a discrete hidden Markov model under stationary and non-stationary situations. 
Our results indicate that even an unsupervised hidden Markov model can outperform LSTM when a massive amount of labeled data is not available. Furthermore, we show that the hidden Markov model can still be an effective method to process the sequence data even when the first-order Markov assumption is not satisfied.

\end{abstract}


\section{Introduction}
\label{sec:headings}
Many real-world applications, such as stock markets, weather, temperature, and DNA sequences, are modeled as time series and sequential problems. The fundamental challenges in time series and sequential data analysis are: 1- Observations at different points in time are correlated. 2-order of observations in time matters. This makes some of the algorithms that change or permute the order of data unusable.

Scientists and researchers have done extensive research in
time series analysis, such as \cite{ariyo2014stock},\cite{tadayon2020comprehensive},\cite{kim2003financial},\cite{zhang2003time}. They have borrowed tools from various domains, such as graphical modeling, statistics, and machine learning to model and forecast the time series data. In \cite{ariyo2014stock}, the authors used autoregressive integrated moving average (ARIMA) to make predictions of the stock market. In \cite{tadayon2020comprehensive}, authors used deep learning, specifically long short term memory (LSTM), to predict multivariate time series data. They showed that even simple LSTM architectures can make an accurate prediction of future values. In [3], the author combined ANN and ARIMA to design a hybrid methodology to model time series data. In [4], the author used machine learning, specifically a support vector machine (SVM), to predict the stock market index.

Graphical models, specifically Bayesian networks (BN), have been used extensively in modeling various applications, such as \cite{hossain2019framework} and \cite{sun2019bayesian}. A dynamic Bayesian network (DBN) specifically an HMM is a variant of a BN that is used to model time series and sequential data in various applications such as \cite{tadayon2020predicting}, \cite{bunian2017modeling}, \cite{rabiner1989tutorial}, \cite{xie2002structure}. In \cite{tadayon2020predicting}, the authors used an HMM to predict student performance in an educational video game. They used a discrete HMM to measure student mastery of concepts as they go through levels of the game. In \cite{bunian2017modeling}, the authors used a five-state HMM to analyze the individual differences in in-game behavior and used the logistic regression for the prediction. In \cite{xie2002structure} the authors incorporated a two-state HMM along with dynamic programming to classify and segment a soccer video
game.

Deep learning has been used extensively in various applications. For instance, LSTM, a variant of the recurrent neural network, has been successful in modeling the sequence data in recent years. For example it is widely used in the natural language processing (NLP) to model sequences \cite{ghosh2016contextual}, \cite{prakash2016neural}. Despite their effectiveness and their numerous applications, neural networks, in general, suffer from the following problems: 1- They require a massive volume of (labeled) data for training. 2-The number of trainable parameters, even for a simple model, is huge. 3- Since they have a huge number of parameters, the models are not easily interpretable. For these reasons, simpler models with fewer parameters are usually preferred if the performance degradation is negligible.

% In \cite{ghosh2016contextual} the authors presented CLSTM (Contextual LSTM), an extension of LSTM to incorporate contextual features (e.g., topics) into the model. In \cite{prakash2016neural} the authors proposed a stacked
% residual LSTM network for efficient training of deep LSTMs
% Another problem with neural network based approaches is that they are inherently data-driven techniques. This means they learn the patterns in data through  iterative method.  
Advances in statistics and optimization allow researchers to develop various algorithms from graph theory and Bayesian learning such as \cite{blei2017variational}, \cite{hoffman2013stochastic}, \cite{jensen1994optimal}, \cite{murphy2013loopy} to design sophisticated graphical models for better inference and learning. However, the more complex the model, the more computationally difficult inference and learning become. HMM and its variants such as \cite{ghahramani1996factorial}, \cite{bengio1995input} have been successfully used in applications with latent variables. This is due to the tree structure of its graph, which makes the inference and learning very efficient.
To the best of our knowledge, this is the first work
that compares LSTM and HMM in terms of the number of samples needed, prediction accuracy, and complexity of models on datasets with various complex patterns.

In this paper, we compare HMM and LSTM in terms of performance, the number of trainable parameters, and how much data is needed for training using synthetic data corresponding to various graph structures. The contribution of this paper is twofold: 1- Comparing LSTM to both supervised and unsupervised HMM in terms of performance and complexity. 2- Proposing some methods to efficiently convert continuous HMM to discrete HMM by discretizing the observations. 

The rest of this paper is organized as follows. Section 2
reviews algorithms and background materials used in other sections. Section 3 describes the data generation mechanism. Section 4 describes the problem formulation. Section 5 presents and discusses the results. Section 6 concludes the paper.

\section{Algorithms}
\label{sec:Algorithm}
\subsection{HMM Algorithms}
In this section, HMM algorithms are briefly reviewed. Both the discrete hidden Markov model (DHMM) as well as the continuous hidden Markov model (CHMM) are discussed.

The HMM is the extension of the Markov process in which the observations are a probabilistic function of the states. In an HMM, states are considered as hidden and should be inferred by the sequence of observations. The HMM is characterized by the following: 

N: Number of the hidden states. Although this is unknown since the states are hidden, it usually can be initialized to a reasonable number depending on the problem and the dataset and later can be learned using various statistical analysis tools which will be discussed later. 

M: Number of the observation symbols per state. 

$\overrightarrow{S}$: State sequence where $\overrightarrow{S}=(s_1,s_2,...s_T)$, T is the length of the sequence, and each $s_i\in \{1,2,...,N\}$.

$\overrightarrow{O}$: sequence of the observation symbols where $\overrightarrow{O}=(o_1,o_2,...o_T)$ and each $o_i\in \{1,2,...,M\}$.

A: State transition probability. It defines the probability of going from state i to the state j and is denoted by
\begin{align}
a_{ij}=p(s_{t+1}=j|s_t=i)
\end{align}

B: Observation distribution per each state, which is denoted as follows:
\begin{align}
b_{i}(k)=p(o_{t}=k|s_t=i)
\end{align}

$\overrightarrow{\pi}$: Initial state distribution that is defined as follows:
\begin{align}
\overrightarrow{\pi_i}=p(s_1=i)
\end{align}

$\lambda$: HMM parameters together are usually denoted by the following:
\begin{align}
\lambda=(A,B,\overrightarrow{\pi})
\end{align}
The above equations together can be used to fully define any HMM with discrete observations. 

Forward and backward algorithms \cite{rabiner1989tutorial} are used to calculate $P(\overrightarrow{O}|\lambda)$, the probability of observing a sequence given $\lambda$. If the time series is not labeled and the mapping between the observations and the states is not available, then HMM parameters should be estimated using the Baum-Welch or EM algorithm \cite{dempster1977maximum}. If the observations are continuous (CHMM) as opposed to discrete, the emission probability distribution should be adjusted to account for this change. Continuous observations are modeled by fitting the probability density functions (pdf) to the data. A Gaussian distribution or mixture of Gaussian distributions are typically used for modeling the data. 

If the observations for each state can be modeled using a single Gaussian distribution, then equation (2) will be changed to the following:
\begin{align}
b_i(x)=p(x|s_t=i)=\textit{N}(x;\mu_i,\Sigma_i)
\end{align}
In equation (5), $\mu_i$ and $\Sigma_i$ are the mean and the covariance matrix of the Gaussian distribution for state i respectively.
     
If a single distribution is not a reasonable fit to the data, then a mixture of Gaussian distributions can be used to model the observations. In this case, equation (6) can be used to model the observations for each state.
\begin{align}
b_i(x)=p(x|s_t=i)=\sum_{m=1}^{M}c_{im}\textit{N}(x;\mu_{im},\Sigma_{im})
\end{align}
$c_{im}$ is the mixture coefficient and determines the weight each component has in modeling the data. $\mu_{im}$ and $\Sigma_{im}$ are the mean and covariance matrices of each mixture component corresponding to the state i. 

Decoding the optimal state sequence given the observation can be done using the Viterbi algorithm \cite{forney1973viterbi}. It finds the sequence of the states that best explains the observed data: 
\begin{align}
S^*=\textit{argmax}_S P(S|O,\lambda)
\end{align}

\subsection{LSTM Algorithm}

RNN networks are mainly designed for the sequence prediction problem. Unlike feed-forward networks that consider inputs and outputs to be independent, RNN networks consist of memory cells that can remember the long term dependencies between elements of a sequence. In theory, RNN can remember arbitrarily long time steps, but in practice, they suffer from the vanishing gradient problem \cite{hochreiter1998vanishing}, \cite{pascanu2013difficulty}. LSTM is designed to address the vanishing gradient problem. The first step in the operation of LSTM is the use of a cell state, which is a memory element to store information. LSTM has the ability to remove and update the cell state via various gates. A forget gate decides what information should be kept or removed. This is done using a sigmoid function that produces a number between 0 and 1, where one means keeping and zero means removing the data. The input gate is responsible for adding or updating the information to the cell state. It consists of a component that involves a sigmoid function to determine which information should be added to the cell state. It also has a component using a tanh to squeeze the data to -1 and 1 range. The output gate decides what information from the cell state should be passed to the output at time t. This is done similarly to the input gate in which the cell state data is squeezed to -1 and 1 range and then is multiplied by the output of a sigmoid function to determine the useful information passed to the output.   
\subsection{K-means Algorithm}

K-means is a well-known distance-based clustering method that uses the Euclidean distance to measure the similarity between data points. The inherent problem with the K-means is that the value of K should be specified in advance. This can be done by the domain knowledge or techniques such as the elbow method, Silhouette index, and gap statistics. 

The elbow method works by iterating through different values of K and plots the sum of squared distances of samples to their closest cluster center. This quantity is called inertia or the sum of squared errors (SSE). Then the optimal number of clusters is selected by the value of K at the “elbow”, namely the point after which the SSE/inertia starts decreasing linearly.

Silhouette index is a measure of how similar a data point is to the points in its cluster compared to points in other clusters. It works by calculating the Silhouette coefficient for each instance of clustering as follows:
\begin{align}
 c= \dfrac{a-b}{max(a,b)}
\end{align}

In equation (8), \textit{a} represents the mean intra-cluster distance, and \textit{b} represents the mean nearest-cluster distance for each sample.
Equation (8) shows that the silhouette coefficient is a number between -1 and 1 in which the number closer to 1 represents that a sample is assigned to the correct cluster and a number closer to -1 shows that the sample is assigned to the wrong cluster.

\section{Dataset}
In this paper, We create the synthetic data corresponding to different DBN structures. We proposed three DBN structures with various levels of complexity.
The followings are the parameters and notations for the synthetic data:

$T$ = Length of each time series.

$D$ = Dimension of time series.

$N$ = Number of samples (time series)

$S_{1},S_{2},...,S_{T}$ = sequence of discrete states.

$u_{i1},u_{i2},...u_{iT}$ = sequence of input i.

$O_{i1},O_{i2},...O_{iT}$ = sequence of observations i.

$N{ui}$ = Number of symbols for input i (if it is discrete).

$N{s}$ = Number of symbols for state i (if it is discrete).

Case I: we generate a structure shown in figure 2 with two confounding variables (inputs) and two continuous observations per time point. For this case, $N$= 2000, $T$=50, $D$=2, $N_{s}$=3, $N_{u1}$ and $N_{u2}$ are both 2.

Case II: we generate a structure shown in figure 3 with one input variable and one continuous observation per time point. This is more complex than the case I since inputs are connected across time in addition to states. Moreover, the input is connected to observation and state per time point. For this case, $N$=2000, $T$=20, $D$=1, $N_{s}$=4, $N_{u1}$ =2.

Case III: we generate a structure shown in figure 4 with one input variable and one continuous observation per time point. This is the most complex case since input at time $t$ is connected to state at time $t$ and input and state at time $t$+1. Moreover, the state at time $t$ is connected to state at time $t$+1 and both observations at time $t$ and $t$+1.For this case, $N$=1000, $T$=10, $D$=1, $N_{s}$=4, $N_{u1}$ =2.

Case IV: we generate a structure shown in figure 5 with one input variable and one continuous observation per time point. This is a more complex version of case II since state at time $t$ is connected to state at time $t$+1 and state at time $t$+2.For this case, $N$=2000, $T$=20, $D$=1, $N_{s}$=4, $N_{u1}$ =2.

For all these cases, a multinomial distribution is used to model discrete nodes, and a Gaussian distribution is used to model continuous nodes. 

This method of generating time series has the advantage that parameters of the time series such as the dimension, length, number of samples, distribution of each node, and graph structures can be arbitrarily chosen. 

\section{Problem Formulation}

In this section, the problem formulation and the prediction
algorithms are discussed. 
The procedure to train and test the LSTM is as follows:
\begin{enumerate}
    \item Pick a training ratio as the amount of data used for training.
    \item Perform transformation on data to scale it to [0,1] range.
    \item Perform model selection to find the optimal parameters.
    \item Test the model on the test (unseen) data. 
    \item Keep reducing the training ratio and perform steps 1 to 4 again.
\end{enumerate}

Figure 1 shows the architectures used for LSTM training:
\begin{figure}[htb]
\centering
\begin{subfigure}{0.5\linewidth}
  \centering
  \includegraphics[width=.4\linewidth]{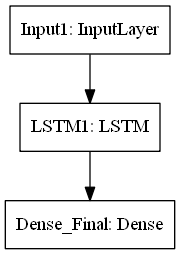}
  \caption{Model 1}
  \label{fig:Img1}
\end{subfigure}%
\begin{subfigure}{.4\linewidth}
  \centering
  \includegraphics[width=.5\linewidth]{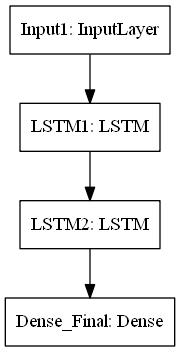}
  \caption{Model2}
  \label{fig:im2}
\end{subfigure}

\begin{subfigure}{.5\textwidth}
  \centering
  \includegraphics[width=.9\linewidth]{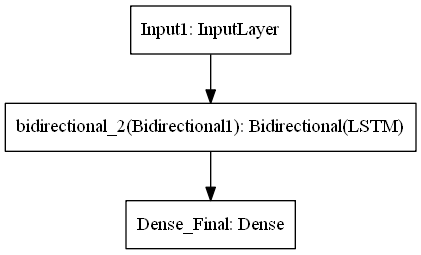}
  \caption{Model3}
  \label{fig:im3}
\end{subfigure}
\caption{LSTM Architectures for Time Series Data}
\label{fig:Dynamic}
\end{figure}

The procedure to train and test the HMM is as follows:
\begin{enumerate}
    \item Pick a training ratio as the amount of data is used for training.
    \item Perform model selection to find the optimal parameters.
    \item Train model with parameters from step 2.
    \item Perform the testing by applying the Viterbi algorithm on test data.
    \item Perform state mapping 
    \item Keep Reducing the training ratio and perform steps 1 to 5 again.
\end{enumerate}

The above procedure works very well for the HMM if there is one feature but if there are more than one feature then a more reasonable method to approach the prediction problem is as follows. First, find the correlation between the features. Second, train one HMM for each set of features that have positive correlation with each other. Third, combine the predictions together.

The differences between DHMM and CHMM lie in the input and how data is modeled.
CHMM can be converted to the DHMM problem by discretizing the observations. The critical question is how to perform discretization in order not to degrade the predictive performance. In this paper, we propose the following:
\begin{enumerate}
    \item \textbf{Network structure is the same through time}:
Since the network structure is unchanged throughout time and number of nodes, their distributions, and confounding variables remain the same. Then time series for different samples can be concatenated to create a long vector, and discretization by domain knowledge or K-means can be performed on this vector.
    \item \textbf{Network structure is varying through time}:
In this case, we perform the discretization per time slice. This is because node distribution,  interconnections, and confounding variables are changing from time to time, and it would not be meaningful to concatenate the time series. For example, consider that the observation corresponds to how long it takes to finish the levels of a game. Different levels have different difficulties, game mechanics, and narratives. Therefore, if some levels are more straightforward than others, then they are expected to take less time, which should be taken into consideration during discretization.
\end{enumerate}

The first case above is used more often than the second one in modeling time series and sequential data since it reduces the inference and learning complexities. This is because it allows us to use dynamic programming to find the estimate of the parameters iteratively while if a network structure is changing with time, a new set of equations is needed to update and estimate the parameters.

This paper aims to measure the trade-off between predictive accuracy and the complexity of HMM and LSTM. We use CHMM for unsupervised HMM, DHMM for both the supervised HMM and unsupervised HMM.K-means clustering is also used for the discretization.

We find the number of trainable parameters for LSTM using simulation. The following are the equations to find the number of parameters in CHMM with diagonal covariance matrix (although the procedure for other cases is very similar) and DHMM. 
\begin{itemize}
    \item CHMM with a diagonal covariance matrix:
    \begin{align}
            C=M*(M-1)+ M*(N-1)+(N*M)*D+(M*N*D)+M-1
    \end{align}
where $M$ is the number of states, $N$ is the number of mixture components, and $D$ is the dimension of time series.
  \item DHMM:
    \begin{align}
            C=M*(M-1)+ M*(N-1)+(M-1)
    \end{align}
where $M$ is number of states, $N$ is number of symbols.
\end{itemize}

It is worth noting that to measure the accuracy of unsupervised HMM (parameters learned from EM), we set the number of states to be $N{s}$ and find the optimal number of mixture components. 

\section{Results and Discussions}
In this section, we present and discuss the results for the comparison of HMM and LSTM.

\textbf{Case I}

This case corresponds to figure 2 and consists of two continuous observations. This can be a simple model of student learning, in which  $S_{1}, S_{2},..., S_{T}$  corresponds to student mastery levels and $u_{1j1}$, $j=1,2,3,...T$ corresponds to study habit (it has two levels corresponding to bad and good) and $u_{2j}$, $j=1,2,3,...T$ corresponds to other factors (like the personal issues, interest,...) and observations can correspond to time to finish an assignment and the assignment score. Tables 1 and 2 summarize the prediction accuracy and the number of parameters for the LSTM and HMM as a function of the training ratio. It is worth noting that $N$=2000 is the total number of time series. Therefore training ratio of 0.8 results in 1600 time series for training and 400 time series for testing. The length of each time series is 50. To train the LSTM model, we usually concatenate the samples, which means we will have 50*1600=80000 samples.
In Table 1 for training ratio=0.8,0.3,0.1, the optimal parameters are as follows: the number of neurons in the hidden layer is 8, the optimizer is Adam, and the batch size is 4. The optimal parameters for training ratio =0.005 and 0.001 are the same, except the number of neurons in the hidden layer is 32. As Table I demonstrates the prediction accuracy for unsupervised HMM even for two time series (2*50=100) samples is very accurate. Both DHMM and CHMM are outperforming LSTM when the number of samples is low. HMM has a significantly lower number of parameters than a simple one layer LSTM.

The correlation between the two observations is -0.87, which shows they are negatively correlated. Therefore two different HMMs will be trained, one for each set of observations, and the final prediction will be a combination of both of them. This method has an advantage that predictions can be weighted and assign more weights to more important observations. 

The unsupervised DHMM provides accuracy comparable to the unsupervised CHMM, which shows that if the discretization is done correctly, there is not much information loss. 

We observe that although under case I, there are two confounding variables per time step, Markov's assumption for states hold and observations are conditionally independent. To make enough samples for HMM and LSTM prediction, instead of increasing the number of cases ($N$), we can increase the length of time series ($T$). This is a crucial observation since, in many domains such as education, it is hard to have a large number of time series and sequence data, but it is rather easier to increase the length of time series. Here we generate synthetic time series, and so we have access to labels (states); therefore, we should manually set the number of states and find the optimal number of mixture components, which is less optimal to perform model selection over states and mixture components. Finally, if discretizing is done correctly, i.e., finding the optimal number of clusters, this can be used as an alternative solution to fit a Gaussian or mixture of Gaussian distributions to data.
\begin{figure}[htb]
    \centering
    \includegraphics{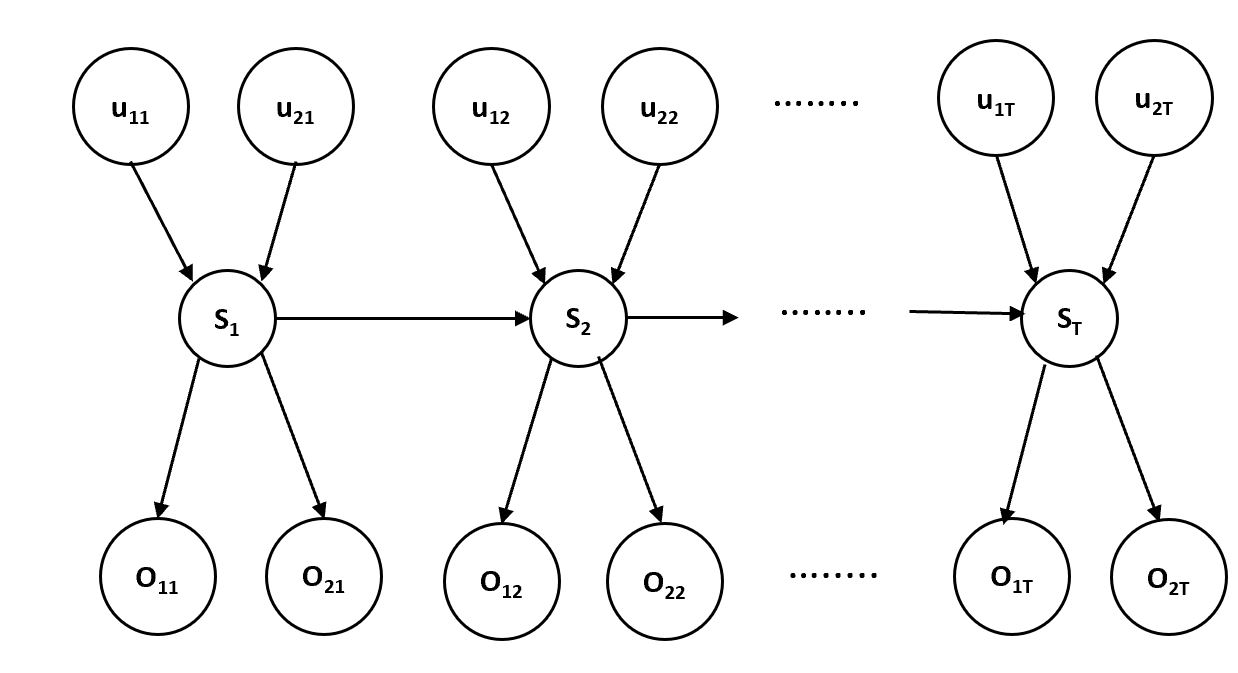}
    \caption{Synthetic Time Series Under Case I}
    \label{fig:label2}
\end{figure}

\begin{table}[htb]
\centering
\caption{LSTM Prediction Accuracy and Number of Parameters for Case I}
\resizebox{0.8\textwidth}{!}{
\begin{tabular}{ |c|c|c|c|c|c| } 
\hline
Architecture & Number of Parameters& Training Ratio & Number of Training Sample &  Accuracy(\%) \\
\hline
\multirow{4}{4em}{Model 1} & \multirow{4}{4em}{379} & 0.8 & 80000 & 98.55 \\
& & 0.3 & 30000 & 98.34\\
& & 0.1 & 10000 & 98.1\\ \cline{2-5}
&\multirow{2}{4em}{4579} & 0.005 & 500 & 93.63\\
& & 0.001 & 100 & 64.2 \\
\hline
\multirow{4}{4em}{Model 2} & \multirow{4}{4em}{971}& 0.8 & 80000 & 98.4 \\
& & 0.3 & 30000 & 98.12\\
& & 0.1 & 10000 & 98.06\\\cline{2-5}
& \multirow{2}{4em}{12899} & 0.005 & 500 & 93.83\\
& & 0.001 & 100 & 69.5 \\
\hline
\multirow{4}{4em}{Model 3} & \multirow{4}{4em}{755}& 0.8 & 80000 & 98.25 \\
& & 0.3 & 30000 & 98.06\\
& & 0.1 & 10000 & 98.06\\\cline{2-5}
&\multirow{2}{4em}{9155} & 0.005 & 500 & 94.23\\
& & 0.001 & 100 & 76.68 \\
\hline
\end{tabular}
}
\end{table}

\begin{table}[htb]
\centering
\caption{HMM Prediction Accuracy and Number of Parameters for Case I}
\resizebox{0.8\textwidth}{!}{
\begin{tabular}{ |c|c|c|c|c| } 
\hline
Model Type& Number of Parameters& Training Ratio & Number of Training Sample &  Accuracy(\%) \\
\hline
\multirow{4}{6em}{Unsupervised CHMM} & 46 & 0.8 & 80000 & 93.89 \\ \cline{2-5}
  &     \multirow{3}{1.3em}{28}        & 0.3 & 30000 & 92.24\\
& & 0.1 & 10000 & 88.50\\
& & 0.005 & 500 & 84.01\\
& & 0.001 & 100 & 81.23 \\
\hline
\multirow{4}{6em}{Unsupervised DHMM} &\multirow{4}{1.5em}{28} & 0.8 & 80000 & 90.89 \\ 
& & 0.3 & 30000 & 87.81\\
& & 0.1 & 10000 & 86.51\\
& & 0.005 & 500 & 78.32\\
& & 0.001 & 100 & 67.01 \\
\hline
\multirow{4}{6em}{Supervised DHMM} & \multirow{4}{1.5em}{28}& 0.8 & 80000 & 96.78 \\
& & 0.3 & 30000 & 96.39\\
& & 0.1 & 10000 & 96.12\\
& & 0.005 & 500 & 96.12\\
& & 0.001 & 100 & 96.12 \\
\hline
\end{tabular}
}
\end{table}

\clearpage

\textbf{Case II}

This case corresponds to the figure 3 and is more complex than case 1 in terms of interconnection between the nodes. In this network, the Markov assumption does not hold since
\begin{align}
    S_{n+1} \not\!\perp\!\!\!\perp S_{n-1} | S_{n}
\end{align}
The prediction procedure is similar to the case I and is summarized in Tables 3 and 4. As Tables 3 and 4 show, the prediction accuracy for this case, especially for unsupervised HMM, is lower than the case I, and this is because the Markov assumption no longer holds for this network. However, unsupervised HMM still provides a reasonable estimate and can outperform LSTM when there are only 200 and 40 samples. The supervised HMM is done by first discretizing the observations according to K-means to five clusters. The elbow and Silhouette methods are used to verify the optimal number of clusters. Comparing the number of parameters between different models suggests that the HMM is much more efficient and less complex than a simple one layer LSTM. Moreover, the sub-optimal unsupervised HMM provides a reasonable prediction even when the structure is not the tree, and the Markov assumption is not satisfied. As Table 4 shows again there is not much loss in terms of prediction accuracy after discretizing the observation. 
\begin{figure}[htb]
    \centering
    \includegraphics{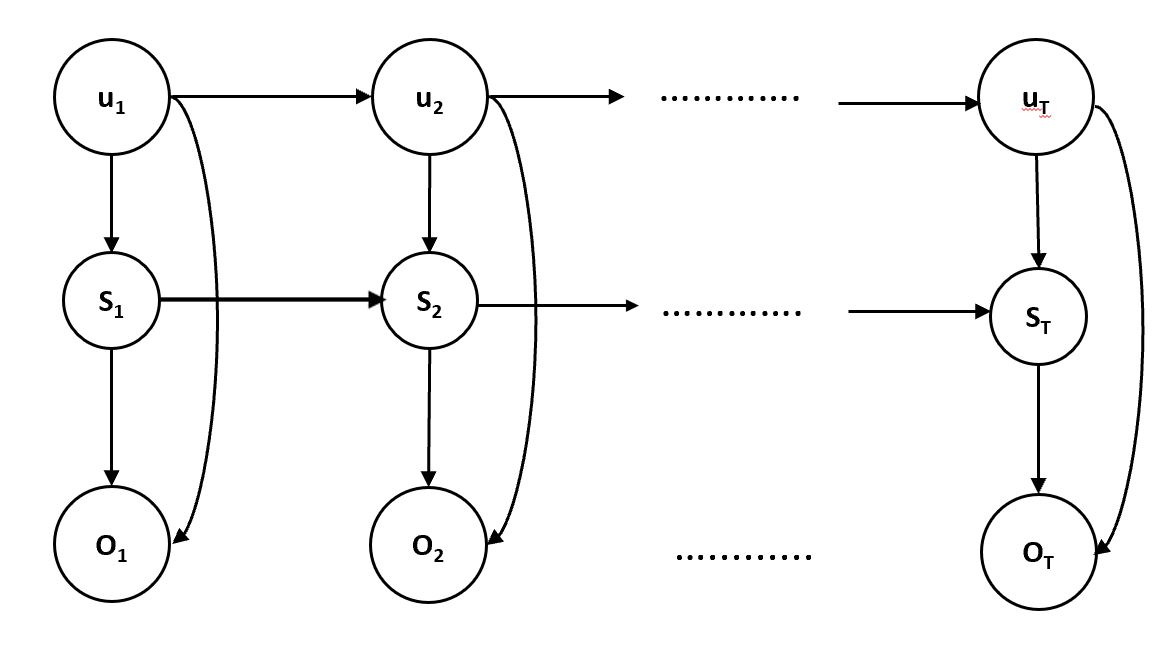}
    \caption{Synthetic Time Series Under Case II}
    \label{fig:label3}
\end{figure}

\begin{table}[htb]
\centering
\caption{LSTM Prediction Accuracy and Number of Parameters for Case II}
\resizebox{0.8\textwidth}{!}{
\begin{tabular}{ |c|c|c|c|c|c| } 
\hline
Architecture & Number of Parameters& Training Ratio & Number of Training Sample &  Accuracy(\%) \\
\hline
\multirow{4}{4em}{Model 1} & \multirow{4}{4em}{356} & 0.8 & 32000 & 88.75 \\
& & 0.3 & 12000 & 87.33\\
& & 0.1 & 4000 & 87.20\\ \cline{2-5}
&\multirow{2}{4em}{4484} & 0.005 & 200 & 58.28\\
& & 0.001 & 40 & 38.23 \\
\hline
\multirow{4}{4em}{Model 2} & \multirow{4}{4em}{900}& 0.8 & 32000 & 88.52 \\
& & 0.3 & 12000 & 88.36\\
& & 0.1 & 4000 & 87.32\\\cline{2-5}
& \multirow{2}{4em}{12804} & 0.005 & 200 & 58.47\\
& & 0.001 & 40 & 33.46 \\
\hline
\multirow{4}{4em}{Model 3} & \multirow{4}{4em}{708}& 0.8 & 32000 & 87.62 \\
& & 0.3 & 12000 & 87.23\\
& & 0.1 & 4000 & 87.19\\\cline{2-5}
&\multirow{2}{4em}{8964} & 0.005 & 200 & 62.47\\
& & 0.001 & 40 & 38.12 \\
\hline
\end{tabular}
}
\end{table}

\begin{table}[htb]
\centering
\caption{HMM Prediction Accuracy and Number of Parameters for Case II}
\resizebox{0.8\textwidth}{!}{
\begin{tabular}{ |c|c|c|c|c| } 
\hline
Model Type& Number of Parameters& Training Ratio & Number of Training Sample &  Accuracy(\%) \\
\hline
\multirow{4}{6em}{Unsupervised CHMM} & 47 & 0.8 & 32000 & 75.21 \\ \cline{2-5}
  &     \multirow{3}{1.3em}{23}        & 0.3 & 12000 & 74.5\\
& & 0.1 & 4000 & 72.14\\
& & 0.005 & 200 & 70.04\\
& & 0.001 & 40 & 74.12 \\
\hline
\multirow{4}{6em}{Unsupervised DHMM} &\multirow{4}{1.5em}{27}  & 0.8 & 32000 & 70.07 \\ 
& & 0.3 & 12000 & 71.36\\
& & 0.1 & 4000 & 70.14\\
& & 0.005 & 200 & 62.83\\
& & 0.001 & 40 & 60.15 \\
\hline
\multirow{4}{6em}{Supervised DHMM} & \multirow{4}{1.5em}{31}& 0.8 & 32000 & 88.91 \\
& & 0.3 & 12000 & 88.15\\
& & 0.1 & 4000 & 88.04\\
& & 0.005 & 200 & 85.09\\
& & 0.001 & 40 & 80.05 \\
\hline
\end{tabular}
}
\end{table}

\vspace{4cm}
\textbf{Case III}

This case corresponds to the figure 4 and is much more complex than cases I and II. In this network, the Markov assumption is not satisfied. Tables 5 and 6 summarizes the prediction accuracy and the number of parameters for this case. In this case, N is 1000, and T is 10. Some interesting observations from this case are:1- Lower prediction accuracies for the HMM and LSTM for this case suggest that a more complex model is needed to model time series. 2- Adding more layers to the LSTM does not necessarily increase the prediction accuracy but can significantly increase the model complexities. 3- Unsupervised HMM, even when the number of states is not optimal, can compete with the supervised HMM or LSTM.

\begin{figure}[htb]
    \centering
    \includegraphics{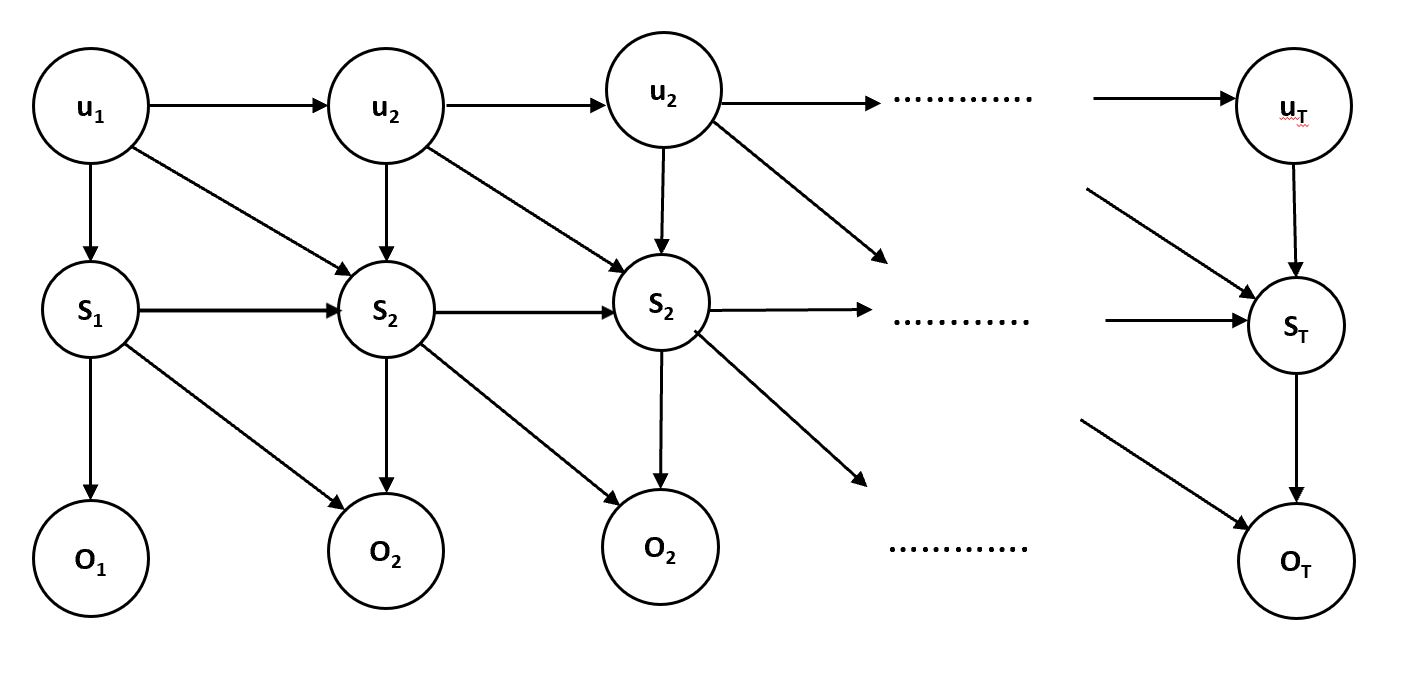}
    \caption{Synthetic Time Series Under Case III}
    \label{fig:label4}
\end{figure}

\begin{table}[htb]
\centering
\caption{LSTM Prediction Accuracy and Number of Parameters for Case III}
\resizebox{0.8\textwidth}{!}{
\begin{tabular}{ |c|c|c|c|c|c| } 
\hline
Architecture & Number of Parameters& Training Ratio & Number of Training Sample &  Accuracy(\%) \\
\hline
\multirow{4}{4em}{Model 1} & \multirow{5}{4em}{4484} & 0.8 & 8000 & 61.59 \\
& & 0.3 & 3000 & 58.36\\
& & 0.1 & 1000 & 56.12\\
& & 0.005 & 50 & 33.84\\
& & 0.001 & 10 & 30.23 \\
\hline
\multirow{4}{4em}{Model 2} & \multirow{4}{4em}{12804}& 0.8 & 8000 & 62.33 \\
& & 0.3 & 3000 & 61.25\\
& & 0.1 & 1000 & 59.51\\
&  & 0.005 & 50 & 34.78\\
& & 0.001 & 10 & 29.10 \\
\hline
\multirow{4}{4em}{Model 3} & \multirow{4}{4em}{8964}& 0.8 & 8000 & 62.32\\
& & 0.3 & 3000 & 61.12\\
& & 0.1 & 1000 & 59.4\\
& & 0.005 & 50 & 34.26\\
& & 0.001 & 10 & 30.12 \\
\hline
\end{tabular}
}
\end{table}

\begin{table}[htb]
\centering
\caption{HMM Prediction Accuracy and Number of Parameters for Case III}
\resizebox{0.8\textwidth}{!}{
\begin{tabular}{ |c|c|c|c|c| } 
\hline
Model Type& Number of Parameters& Training Ratio & Number of Training Sample &  Accuracy(\%) \\
\hline
\multirow{4}{6em}{Unsupervised CHMM} & 35 & 0.8 & 8000 & 60.8 \\ \cline{2-5}
  &     \multirow{3}{1.3em}{23}        & 0.3 & 3000 & 60.15\\
& & 0.1 & 1000 & 58.4\\
& & 0.005 & 50 & 51.74\\
& & 0.001 & 10 & 37.67 \\
\hline
\multirow{4}{6em}{Unsupervised DHMM} &\multirow{4}{1.5em}{31} & 0.8 & 8000 & 60.95 \\ 
& & 0.3 & 3000 & 60.01\\
& & 0.1 & 1000 & 57.9\\
& & 0.005 & 50 & 50.16\\
& & 0.001 & 10 & 37.2 \\
\hline
\multirow{4}{6em}{Supervised DHMM} & \multirow{4}{1.5em}{35}& 0.8 & 8000 & 61.32 \\
& & 0.3 & 3000 & 60.10\\
& & 0.1 & 1000 & 60.10\\
% & & 0.005 & 50 & 85.09\\
% & & 0.001 & 10 & 80.05 \\
\hline
\end{tabular}
}
\end{table}
\clearpage

\textbf{Case IV}
This case corresponds to the figure 5. In this network, the Markov assumption is not satisfied, and state dependencies are over two time steps. Tables 6 and 7 summarizes the prediction accuracy and the number of parameters for this case. In this case, N is 2000, and T is 20. Some observations from this case are:1- DHMM provides comparable accuracy to CHMM if discretization is done correctly. 2- Supervised HMM outperforms LSTM for a lower number of samples even when Markov assumption does not hold, and state dependencies are more than one step time.

Some observations from all these four cases are as follows: 1- HMM, unlike LSTM, can be used as in an unsupervised domain to model sequential and time series data. This is important especially since the amount of labeled time series is limited.2- HMM is a model-based generative model that operates based on the first-order Markov assumption, however LSTM is a data-driven discriminative model that can learn information from arbitrarily long time-steps. 3- HMM has significantly fewer parameters than LSTM, which makes it easier to interpret.4- LSTM is a data-driven method and needs lots of samples for efficient training.5- LSTM has a huge number of hyper-parameters, such as the number of layers, batch-size, choice of the optimization method, learning rate, and the number of neurons per layer. It is very hard, and there is no ground rule to find optimal parameters in LSTM. 6- HMM is very sensitive to the initial condition due to the EM algorithm; therefore, the model might never converge to the global optima. One way to overcome this issue is to start the EM algorithm from multiple different initial conditions and choose the model that optimizes some criteria.
7-Supervised HMM provides a very high prediction accuracy even when Markov assumption fails, and state dependencies are more than one time step.

\begin{figure}[htb]
    \centering
    \includegraphics{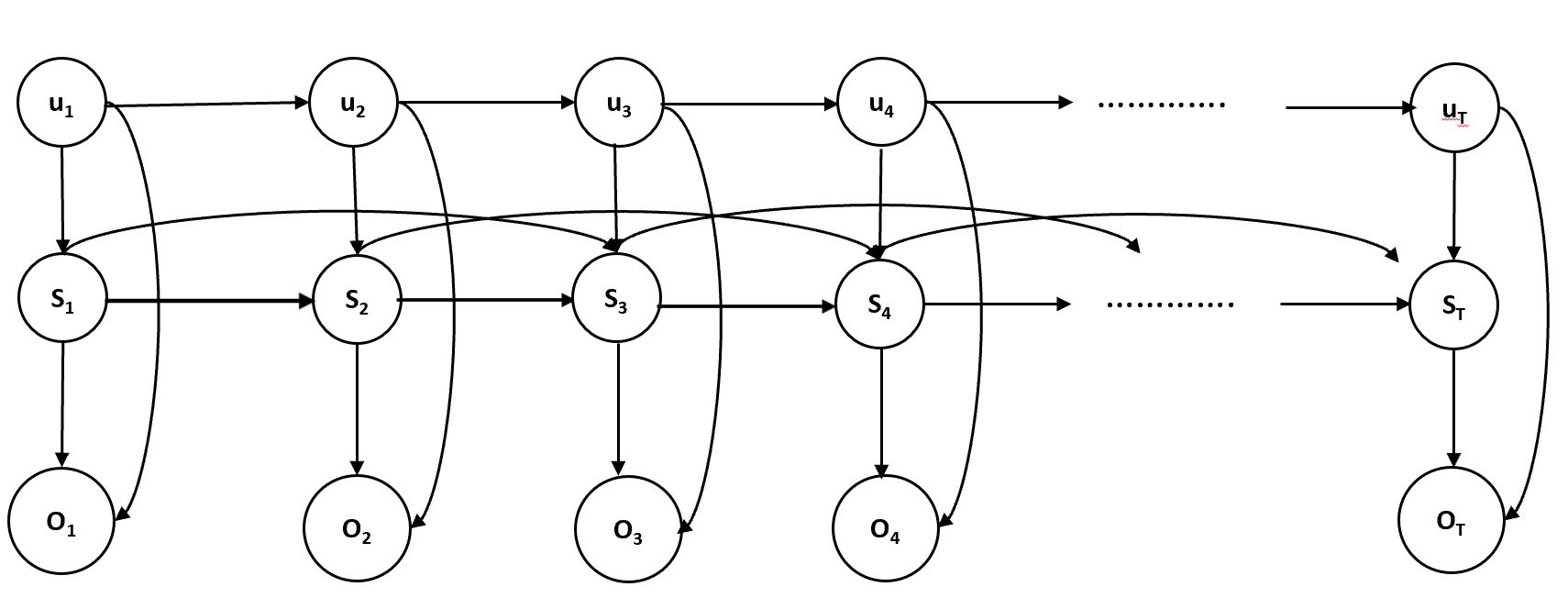}
    \caption{Synthetic Time Series Under Case IV}
    \label{fig:label5}
\end{figure}

\begin{table}[htb]
\centering
\caption{LSTM Prediction Accuracy and Number of Parameters for Case IV}
\resizebox{0.8\textwidth}{!}{
\begin{tabular}{ |c|c|c|c|c|c| } 
\hline
Architecture & Number of Parameters& Training Ratio & Number of Training Sample &  Accuracy(\%) \\
\hline
\multirow{4}{4em}{Model 1} & \multirow{5}{4em}{4484} & 0.8 & 32000 & 82.97 \\
& & 0.3 & 12000 & 81.68\\
& & 0.1 & 4000 & 80.54\\
& & 0.005 & 200 & 36.84\\
& & 0.001 & 40 & 34.33 \\
\hline
\multirow{4}{4em}{Model 2} & \multirow{4}{4em}{12804}& 0.8 & 32000 & 83.30 \\
& & 0.3 & 12000 & 81.92\\
& & 0.1 & 4000 & 80.70\\
&  & 0.005 & 200 & 43.36\\
& & 0.001 & 40 & 35.07 \\
\hline
\multirow{4}{4em}{Model 3} & \multirow{4}{4em}{8964}& 0.8 & 8000 & 82.90 \\
& & 0.3 & 12000 & 81.72\\
& & 0.1 & 4000 & 81.01\\
& & 0.005 & 200 & 40.07\\
& & 0.001 & 40 & 33.89 \\
\hline
\end{tabular}
}
\end{table}

\begin{table}[htb]
\centering
\caption{HMM Prediction Accuracy and Number of Parameters for Case IV}
\resizebox{0.8\textwidth}{!}{
\begin{tabular}{ |c|c|c|c|c| } 
\hline
Model Type& Number of Parameters& Training Ratio & Number of Training Sample &  Accuracy(\%) \\
\hline
\multirow{4}{6em}{Unsupervised CHMM} & 59 & 0.8 & 32000 & 68.4 \\ \cline{2-5}
  &     \multirow{3}{1.3em}{23}        & 0.3 & 12000 & 66.2\\
& & 0.1 & 4000 & 63.47\\
& & 0.005 & 200 & 61.12\\
& & 0.001 & 40 & 58.12 \\
\hline
\multirow{4}{6em}{Unsupervised DHMM} & 27 & 0.8 & 32000 & 64.17 \\ 
& 35 & 0.3 & 12000 & 62.58\\
& 27 & 0.1 & 4000 & 62.40\\
& 35 & 0.005 & 200 & 60.12\\
& 35 & 0.001 & 40 & 57.89 \\
\hline
\multirow{4}{6em}{Supervised DHMM} & \multirow{3}{1.5em}{35}& 0.8 & 32000 & 83.89 \\
& & 0.3 & 12000 & 83.77\\
& & 0.1 & 4000 & 83.11\\
& & 0.005 & 200 & 81.36\\ \cline{2-5}
&23 & 0.001 & 40 & 50.91 \\
\hline
\end{tabular}
}
\end{table}

\section{Conclusion}

In this paper, we created synthetic time series corresponding to various DBN structures with different degrees of complexity. We calculated and compared the prediction accuracies and number of parameters for the LSTM and supervised and unsupervised HMM. We showed that even an unsupervised HMM can be a reliable method when the amount of labeled data is limited. Furthermore, We showed that unsupervised DHMM can provide comparable prediction performance to unsupervised CHMM when observations are continuous. We also showed that a supervised DHMM outperforms LSTM and produces reliable and accurate predictions when the number of samples is limited. We proposed a method to discretize the observation under two different scenarios, i.e.,  1- The network structure is changing with time 2- network structure remains the same. The second option is used in synthetic data generation as it significantly reduces the number of parameters.

\bibliographystyle{unsrt}  
\bibliography{references}  %%% Remove comment to use the external .bib file (using bibtex).

%\begin{thebibliography}{1}

%\bibitem{kour2014real}
%George Kour and Raid Saabne.
% /*\newblock Real-time segmentation of on-line handwritten arabic script.
% \newblock In {\em Frontiers in Handwriting Recognition (ICFHR), 2014 14th
%   International Conference on}, pages 417--422. IEEE, 2014.

% George Kour and Raid Saabne.
% \newblock Fast classification of handwritten on-line arabic characters.
% \newblock In {\em Soft Computing and Pattern Recognition (SoCPaR), 2014 6th
%   International Conference of}, pages 312--318. IEEE, 2014.

% \bibitem{hadash2018estimate}
% Guy Hadash, Einat Kermany, Boaz Carmeli, Ofer Lavi, George Kour, and Alon
% \newblock Estimate and replace: A novel approach to integrating deep neural
%   networks with existing applications.
% \newblock {\em arXiv preprint arXiv:1804.09028}, 2018.

% \end{thebibliography}
% */

\end{document}